\documentclass{article}
\usepackage[utf8]{inputenc}
\title{Tracking Holistic Object Representations}
\usepackage[skip=5pt]{caption}
\usepackage[symbol]{footmisc}

\author{Axel Sauer\footnotemark[1], Elie Aljalbout\footnotemark[1], Sami Haddadin\\
Munich School of Robotics and Machine Intelligence\\
Technical University of Munich\\
\big\{name.lastname\big\}@tum.de}
\date{}

\usepackage{amsmath,amssymb,amsthm}
\usepackage{graphicx}
\usepackage{xspace}
\RequirePackage[dvips]{epsfig} 
\RequirePackage{float}
\RequirePackage{fancyhdr}
\RequirePackage{pstricks,pst-plot,psfrag}
\RequirePackage{multirow}
\RequirePackage{rotating}
\RequirePackage{units}
\usepackage{booktabs}
\usepackage{pgfplots}
\newlength\figureheight
\newlength\figurewidth
\pgfplotsset{compat=1.14}
\usepackage{caption}
\usepackage{array}
\usepackage{subcaption}
\usepackage{setspace}
\usepackage{enumitem}

\newcommand{\ra}[1]{\renewcommand{\arraystretch}{#1}}

\usepackage{tabularx}
\newcolumntype{L}[1]{>{\raggedright\arraybackslash}p{#1}}
\newcolumntype{C}[1]{>{\centering\arraybackslash}p{#1}}
\newcolumntype{R}[1]{>{\raggedleft\arraybackslash}p{#1}}

\usepackage{etoolbox,siunitx}
\usepackage[super]{nth}
\usepackage{nicefrac}
\robustify\bfseries

\newcommand\MyBox[2]{
  \fbox{\lower0.75cm
    \vbox to 1.7cm{\vfil
      \hbox to 1.7cm{\hfil\parbox{1.4cm}{#1}\hfil}
      \vfil}%
  }%
}

\newcommand\MyBoxSmall[2]{
  \fbox{\lower0.0cm
    \vbox to 0.65cm{\vfil
      \hbox to 0.65cm{\hfil\parbox{1.4cm}{#1}\hfil}
      \vfil}%
  }%
}

\makeatletter
\DeclareRobustCommand\onedot{\futurelet\@let@token\@onedot}
\def\@onedot{\ifx\@let@token.\else.\null\fi\xspace}
\def\eg{e.g\onedot}

\makeatother

\begin{document}
\renewcommand*{\thefootnote}{\fnsymbol{footnote}}
\footnotetext[1]{Shared first authorship.}
\renewcommand*{\thefootnote}{\arabic{footnote}}
\maketitle
\begin{abstract}
Recent advances in visual tracking are based on siamese feature extractors and template matching. For this category of trackers, latest research focuses on better feature embeddings and similarity measures. In this work, we focus on building holistic object representations for tracking. We propose a framework that is designed to be used on top of previous trackers without any need for further training of the siamese network. The framework leverages the idea of obtaining additional object templates during the tracking process. Since the number of stored templates is limited, our method only keeps the most diverse ones. We achieve this by providing a new diversity measure in the space of siamese features. The obtained representation contains information beyond the ground truth object location provided to the system. It is then useful for tracking itself but also for further tasks which require a visual understanding of objects. Strong empirical results on tracking benchmarks indicate that our method can improve the performance and robustness of the underlying trackers while barely reducing their speed. In addition, our method is able to match current state-of-the-art results, while using a simpler and older network architecture and running three times faster.
\end{abstract}

\section{Introduction}
Visual tracking is a fundamental computer vision problem, which has been receiving rapidly expanding attention lately. Template-matching methods for tracking are among the most popular ones, due to their fast speed and high accuracy. Briefly, these methods use a template of the target object and try to match it to regions of the image in question. The template usually corresponds to a patch in the first frame~\cite{bertinetto2016fully,tao2016siamese,li2018high,zhu2018distractor}, a previous frame~\cite{held2016learning} or in some cases an interpolation of recently identified patches of the tracked object~\cite{valmadre2017end}. The key for good performance is dependent on the quality of the provided template, as it is the only information given to the tracker. Moreover, the space in which the actual template matching is applied is very influential to the performance of these methods. 
For this reason, several feature extractors have been proposed over the years to improve the performance of tracking algorithms. Prior approaches relied mostly on hand-engineered features to describe the target object~\cite{olson1997automatic,steger2002occlusion,hofhauser2008edge,mohr2009continuous}. In most cases, hand-crafted features are specific to certain tasks and fail to generalize to various scenarios and environmental conditions. Recently, there has been a shift towards using neural networks as feature extractors. Specifically, siamese networks are used to learn a space embedding for the template matching ~\cite{bertinetto2016fully}. Methods based on siamese networks currently dominate most tracking benchmarks ~\cite{VOT_TPAMI, OTB2015}.

\begin{figure}[htb!]
    \centering
    \includegraphics[width=\linewidth]{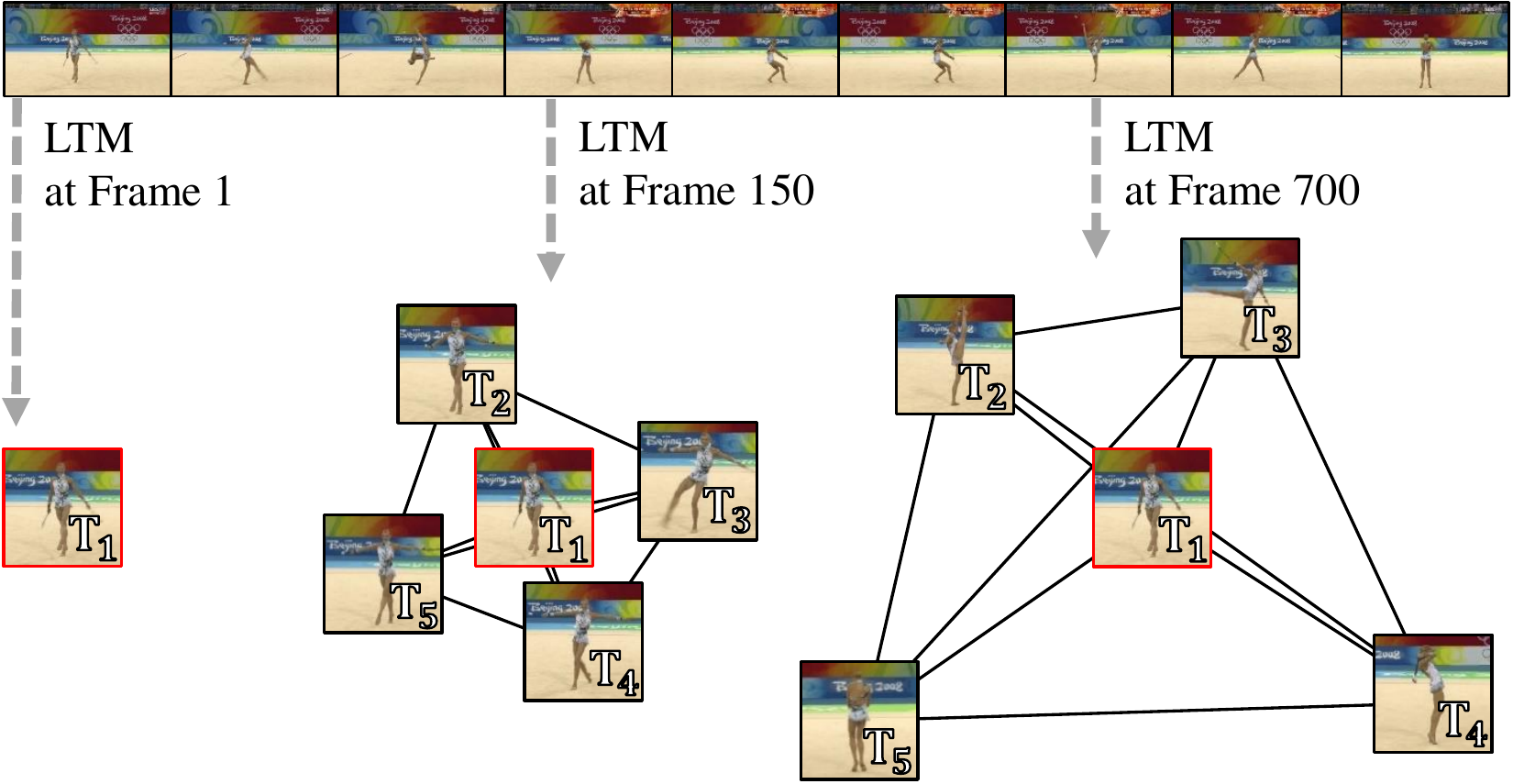}
    \caption{\textbf{Tracking Holistic Object Representations (THOR).} 
    The task in the sequence \textit{gym}, of the tracking benchmark OTB100 ~\cite{OTB2015}, is to track the gymnast. The goal of THOR is to maximize the diversity of the tracked object's representation while tracking. For explanatory purposes, we illustrate this representation, accumulated in the long-term module (LTM), in 2D. In reality, the representation occupies a high dimensional space. Over time, the total volume of the parallelotope spanned by the templates, increases. The base template $T_1$ always stays in the LTM and can be thought of as a fixed point of the parallelotope.}
    \label{fig:templates}
\end{figure}

Nevertheless, independent of the features used for template matching, a crucial aspect of this family of methods is the selection of the template as well as the representation of the template itself. While previous methods focus on finding a good mapping to more matching-friendly spaces for single templates, this paper presents an approach to build template modules representing the variation of the object's appearance in time. Namely, in a dynamic environment, an object can be subject to several condition changes, such as rotation (of the object or the camera), illumination, occlusions, motion blur and even changes in the object shape (\eg{ due to a deformation}). The main goal of this paper is to present a framework, which enables building template modules accounting for all these problems and any other variations that the object could endure during tracking. The presented approach can be considered an extension to \textit{any} template matching based tracker which uses an inner product operation for similarity computation. The main idea of our approach is to find templates that are the furthest away from each other in feature space, as illustrated in Figure \ref{fig:templates}. For every newly introduced tracker based on this principle, our system can always be plugged on top, to increase performance and robustness with only a small set of hyperparameters. Furthermore, our method does not require any (re-)training of the network and barely affects the speed of the original tracker. Our code\footnote{https://github.com/xl-sr/THOR} and videos \footnote{https://sites.google.com/view/vision-thor/} are available online.

\section{Related Work}
Several different approaches have been presented to solve the visual tracking problem. Since our main contribution is to build holistic multi-template modules, we briefly review template matching methods for tracking, in addition to siamese networks based-methods, which combined with our method, perform the best on tracking benchmarks. Moreover, we review approaches which also aim at building holistic object representations.

\subsection{Template-matching}
Briefly, template matching is a computer vision technique for finding areas in an image that match a template image, where matching corresponds to satisfying some similarity constraint. The matching process can either be applied to image intensities (using measures such as normalized cross-correlation or sum of squared differences), gradient features~\cite{mohr2009continuous}, or more generally any relevant features for the tracking problem~\cite{olson1997automatic,steger2002occlusion,hofhauser2008edge}. Recently, with the success of deep neural networks (DNNs) in image recognition tasks, such models have been used as feature extractors for tracking. For instance, siamese instance search tracking (SINT) ~\cite{tao2016siamese} trains a siamese network using the margin contrastive loss for feature extraction. Based on these features, they compare the target image with patches sampled around the previously detected image location. The patch with the highest score is then considered to be matching. Unlike SINT, fully convolutional siamese networks (SiamFC) ~\cite{bertinetto2016fully} uses fully convolutional networks which remove the bias towards the central subwindow of the image. Furthermore, SiamFC uses the embedding of the template as the correlation filter for the search image, which allows real-time performance. As an extension to this approach, SiamRPN~\cite{li2018high} uses region proposal networks~\cite{ren2015faster} to perform proposal extraction on the correlation feature maps. Additionally, they augment the SiamFC architecture with a bounding box regression branch similar to the one used in SINT. To account for class imbalance between positive and negative samples during the training process of SiamRPN, ~\cite{zhu2018distractor} uses distractor objects in previous patches as negative samples at the current ones. This extension improves the features representation to better distinguish target objects from resembling distractors. Unlike previously mentioned methods, GOTURN~\cite{held2016learning} skips the patches sampling step, and only inputs the search image and a patch of the current image (centered around the old detection) and regresses the position of the bounding box. In spite of the advantages of this approach concerning speed and handling aspect ratio and scale changes, the accuracy of this method is inferior to state-of-the-art methods.

\subsection{Multi-template modules}
One major problem in tracking is the change of object appearances. In template matching based methods, this can be solved either by updating the template itself~\cite{nguyen2001occlusion,held2016learning,valmadre2017end}, or by building a representation on top of the template which accounts for this problem. For instance, in~\cite{black1998eigentracking, jurie2002real}, template matching is performed in the eigenspace representation of the template,  which can provide a compact approximate encoding of a large set of images in terms of a small number of orthogonal basis images. More recently,~\cite{yang2018learning} presents an approach for building dynamic memory networks which adapt the template to the target's appearance variations during tracking. Their method uses an LSTM \cite{hochreiter1997long} as a memory controller to read and write templates to their template memory based on memory neural networks~\cite{weston2014memory, sukhbaatar2015end}. Reading, in that case, corresponds to choosing the appropriate template, while writing corresponds to the decision of whether to save an image as a template or not. Based on this approach, their method MemTrack builds a multi-template representation of the object. Another similar work is MMLT \cite{lee2018memory}. This method accumulates feature templates about the object and then uses a weighted combination of the features to convolve over the image. We argue that a mathematical combination of features might only make sense when assuming that all stored templates are of the original object.

Similarly, this paper aims at presenting a new approach for building holistic object representations. Unlike MemTrack, we build our representations in an analytical way rather than embedding this problem into the learning process. Besides the resulting speed advantage, the analytical approach introduces more interpretability to both the obtained representation and the corresponding building process. Additionally, this allows our method to be added to any template matching based tracker with no need for additional training, while MemTrack requires simultaneous training of the memory networks and the trackers using sequential data. In contrast to MMLT, we use all stored templates individually rather than combining them. The main reason for this is that we only store templates which are diverse enough to represent the object but also similar enough to the base template to avoid drifting to distractor objects. To do so, we present both a diversity measure of the stored templates as well as lower bound on the similarity between candidate and base templates.

\begin{figure}[htb!]
    \centering
    \includegraphics[width=1.0\linewidth]{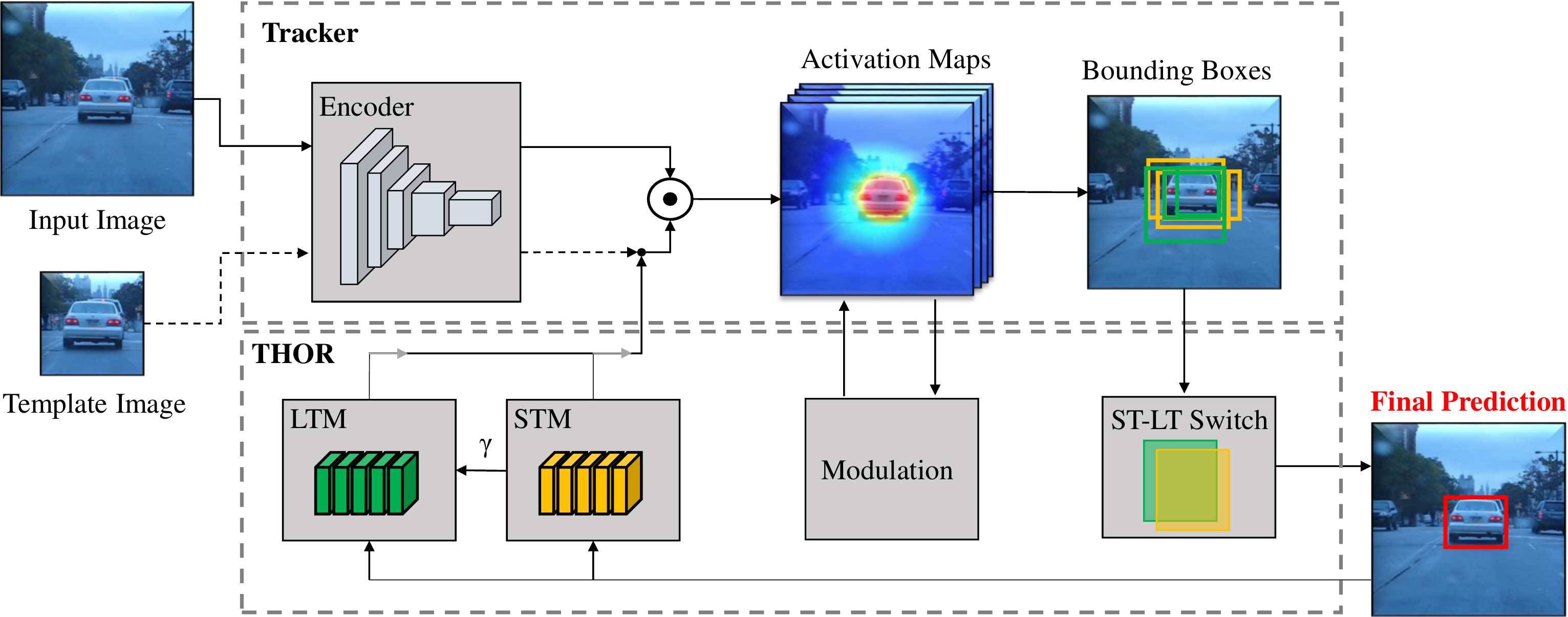}
    \caption{\textbf{System Overview.} The tracker and THOR can be considered separate components that exchange information.
    The \textit{input image} and the initial \textit{template image} are passed through an \textit{encoder} (the template image only at the beginning of the sequence), transforming both into feature vectors in an inner product space. The activation maps are then computed with a dot product. For siamese trackers, the encoder is a siamese network and the dot product is a convolution. Over time, THOR accumulates long-term (LT) and short-term (ST) templates. Convolving the accumulated templates with the input image yields two sets of activations maps (corresponding to LT and ST templates). The \textit{modulation} module calculates a weighted spatial average and multiplies it with all activation maps. Based on these activation maps, the tracker computes the bounding boxes. The box with the highest score in each set is fed into the \textit{ST-LT switch} which determines which bounding box to use for the prediction. The final prediction is then fed back to the \textit{STM} and \textit{LTM} modules to decide whether to keep it or not. The STM also passes the diversity measure $\gamma$ to the LTM.}
    \label{fig:system}
\end{figure}

\section{THOR: Tracking Holistic Object Representations} \label{sec:approach}

In order to build a rich multi-template representation for tracking, we present a framework composed of a \textbf{long-term module (LTM)} and a \textbf{short-term module (STM)}. The LTM represents the tracked objects in diverse conditions (lighting, shape, etc.). It is used to track and re-detect the object in the long term. The STM  selects templates representing short-term variations of the object's appearance. The full system is shown in Figure~\ref{fig:system}. The idea of using both long-term and short-term features for tracking have previously been exploited \cite{hong2015multi, lee2018memory}. However, THOR is different from previous methods in the way features are found and used for tracking.

\subsection{Long-term Module}\label{sec:longtermmodule}

The goal of this module is to store templates which maximize the diversity of information about the tracked object. A na\"ive approach would be to store all tracked crops of an object. Such an approach would be intractable in terms of memory and computational expense. Therefore, only a limited number of templates $K_{lt}$ can be kept. Thus, a crop should only be stored as a template if it contains additional information about the object compared to the already accumulated templates.

\textbf{Problem Statement.} The goal is to find templates during tracking which are the most diverse and furthest away from each other concerning the information they contain about the object. In the real world, an object's state is affected by several object-specific properties such as material type, colour, shape and the dynamics of the object, but also by environmental properties such as illumination, temperature, etc. Only a certain amount of this information is recoverable from a 2D image of an object, which makes an exhaustive object description impossible. A practical alternative to this problem is to describe the object's state with visual features. In this work, we focus on visual features extracted from Siamese networks. The space of n-dimensional visual features together with the convolution operator form an inner product space, which enables us to efficiently compute a diversity measure.

\textbf{Allocation Strategy.} Mathematically, the goal of the LTM is to maximize the volume $\Gamma(f_1,\dots,f_n)$ of the parallelotope formed by the feature vectors $f_i$ of the template $T_i$. In deep template matchers, the feature vector of the given template image $T_1$ is treated as a convolutional kernel. The siamese network embeds images in a feature space where the convolution operator is a measure of similarity \cite{bertinetto2016fully}. During tracking, the template kernel $f_1$ is applied to the input image to get the location of the highest similarity. Hence, if we want to measure how similar two templates $T_1$ and $T_2$ are, we can calculate $f_1\star f_2$. Doing this with all templates in memory, we can construct a Gram matrix:
\begin{gather}
G(f_1,\cdots,f_n)=  
\begin{bmatrix}
    f_1\star f_1 & f_1 \star f_2 &\cdots &f_1 \star f_n \\
    \vdots       &\vdots         &\ddots &\vdots        \\
   f_n \star f_1 &f_n \star f_2  &\cdots &f_n \star f_n
\end{bmatrix}
\end{gather}

$G$ is a square $n\times n$ matrix, where $n$ is typically much smaller than the dimensionality of the feature space. The determinant of $G$, called the Gram determinant, is the square of the $n$-dimensional volume $\Gamma$ of the parallelotope constructed on $f_1, f_2, \dots, f_n$. Therefore, the objective can be written as
\begin{equation}
\underset{f_1,f_2, \dots, f_n}{max} \Gamma(f_1,\dots,f_n) \propto \underset{f_1,f_2, \dots, f_n}{max} |G(f_1,f_2, \dots, f_n)|
\end{equation}
A template is only taken into memory if it increases the Gram determinant of the current LTM when replacing one of the allocated templates. The vectors $f_i$ in this case, can be thought of as basis vectors of the features space, representing the tracked object's manifold in this embedded representation. The maximum number of templates in this framework is the dimensionality of the feature space $D$ (while ignoring memory limitations). For $n > D$, the determinant would be zero. In the practical setting we have  $n \ll D$.

\textbf{Lower bound.}
Candidate templates for the long-term module are bounding boxes detected by the tracker. Hence, the store could end up containing irrelevant images due to tracking drift. In such cases, using templates from the long-term store could push the tracker to drift towards different objects in the scene and the long-term store would continue to deteriorate. To avoid this problem, one could set an upper bound on $|G|$. However, since finding such a value is not straight-forward, we propose to use a lower bound on the similarity measure between a candidate template $T_c$ and the base template $T_1$. $T_1$ is the only ground truth available to the tracker, therefore a new template needs to satisfy $f_c \star f_1 >  \ell \cdot G_{11}$. The parameter $\ell$ can be seen as a temperature on the similarity of $T_1$ on itself and can be used to trade-off tracking performance against robustness against drift. 

In many cases, however, a static boundary on the base template $T_1$ is too conservative. We, therefore, propose two strategies: (i) A \textit{dynamic} lower bound. To take the short-term change of appearance into account we subtract a diversity measure $\gamma$ given by the STM. A valid template is found if $f_c \star f_1 >  \ell \cdot G_{11}  - \gamma$. (ii) An \textit{ensemble} lower bound. We keep the bound static, however, it is used with respect to all templates in the LT module. This enables much lower values for $\ell$ while still being robust against drift. A valid template satisfies (all) the inequalities in $f_c \star f_{1:n} > \ell \cdot diag(G)$

\textbf{Template Masking.}
Following \cite{bolme2010visual}, we multiply the feature vector $f_i$ with a tapered cosine window before calculating the similarities between templates. This reduces the influence of the background of a template. This transformation changes the space in which the Gram determinant for the LTM is calculated. However, we argue that all computations of the LTM are still consistent in this new space since the same mask is applied to all templates, independent of the background-foreground ratio.

\subsection{Short-term Module}\label{sec:shorttermmodule}
The goal of the STM is to handle abrupt movements and partial occlusion. Both of these cases cannot be handled by the LTM. The templates created in these situations are too dissimilar to the base template and would be rejected by the LTM. The module slots of the STM are allocated in a first-in, first-out manner, the number of slots is set to a fixed number $K_{st}$.

We also leverage the object representation in the STM to calculate a diversity measure $\gamma$. Given the Gram matrix $G_{st}$ of the STM, we could use the Gram determinant to calculate the diversity as we do for the LTM. However, in the short term, this measure is not well behaved, it strongly fluctuates, takes on very small or very large values and we have nothing to normalize it with. Instead, we calculate the diversity measure as 
$$\gamma = 1 - \frac{2}{N(N+1)G_{st, max}} \sum^{N}_{i<j}{G_{st, ij}}$$ 
In words, we sum up the upper triangle of the Gram matrix and normalize the sum by the maximum value in the Gram matrix. This puts $\gamma$ in the range of $[0-1]$, the closer $\gamma$ to $1$, the more diverse the templates in the STM.

\subsection{Inference Strategy.} To get a predicted bounding box, we apply two methods during inference. \textit{Modulation} aims to leverage all of the information that is contained in both STM and LTM. The \textit{ST-LT Switch } determines which template yields the current best prediction and outputs the final bounding box.

\textbf{Modulation.} At every frame, we get the activation maps of every template in both STM and LTM. To use the predictions of all templates, we compute a weighted spatial average over all activation maps.  The weights correspond to the maximum scores for each template, i.e. if a template is more certain it contributes more to the average. Every activation map is then multiplied by this average activation map and re-normalized.

\textbf{ST-LT Switch.} By default, we always use the predicted bounding box of the STM, since it can handle short term challenges well. However, since no template stays in the STM permanently, it is prone to tracking drift. In visual tracking, drift is determined by calculating the intersection over union ($IoU$) of a predicted bounding box and the ground-truth \cite{VOT_TPAMI}. We leverage this measure of drift and calculate the $IoU$ between the two bounding boxes of the STM and LTM with highest scores. In this case, we treat the prediction of the LTM as ground truth since it is more robust against drift. If the $IoU(STM, LTM)$ is lower than a threshold $th_{iou}$, we use the prediction of the LTM and reinitialize the STM.

\subsection{Implementation Details}
To keep the memory updates and the forward pass efficient, we use two strategies: parallelization and dilation. For the memory updates, we need to compute the similarities between the template candidate and all templates in memory. The same also applies for a forward pass, where we need to compute the activation maps for all templates. The operation to compute the similarities is a 2D convolution, this means that we can calculate all similarities in parallel. Therefore, if the GPU memory is large enough, these operations slow down the tracker only slightly, see Section \ref{sec:sota_comparisons}. Moreover, since consecutive frames are very similar in appearance, only every other frame is considered as a template. We set a constant dilation value of $10$, i.e. every tenth frame is fed into the STM and LTM.

\section{Experiments}
We build experiments to answer the following questions: (i) Does the determinant increase throughout a tracking sequence and does it converge? (ii) What is the effect of THOR on the speed of the used trackers? (iii) Does THOR improve the performance of state-of-the-art trackers? (iv) What is the effect of each introduced concept (modulation, masking, lower bound, short-term module) on the performance of our method?

Generally, THOR's underlying principle can be applied to any template matching tracker. For this work we compare the following siamese network based trackers: SiamFC ~\cite{bertinetto2016fully}, SiamRPN ~\cite{li2018high}, and SiamMask ~\cite{wang2018fast}. The tunable parameters of THOR are the number of memory slots in STM $K_{st}$ and LTM $K_{lt}$, the IoU threshold of the ST-LT switch $th_{iou}$, the lower bound $\ell$ and $\alpha$ of the tapered cosine window. We use PyTorch for the implementation and the experiments were done on a PC with an Intel i9 and an Nvidia RTX 2080 GPU. In the following, we give a proof of concept, report the performance of on VOT2018 ~\cite{VOT_TPAMI} and OTB2015 ~\cite{OTB2015}, and conduct an ablation study. 

\subsection{Proof of Concept} \label{sec:poc}
\begin{figure}
\centering
   \begin{minipage}{0.37\linewidth}
       \includegraphics[width=\textwidth]{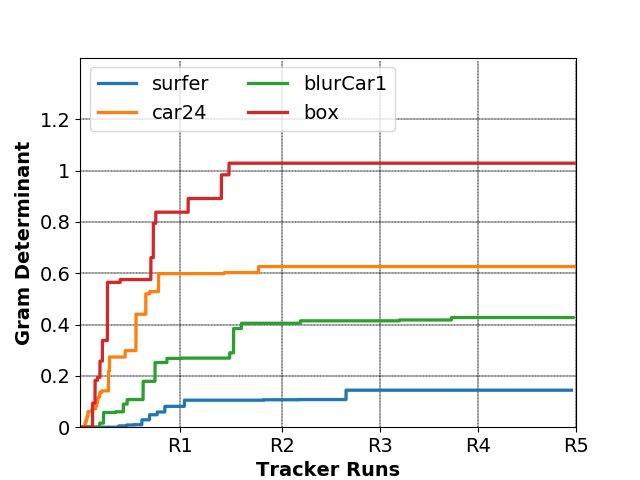}
     \end{minipage}%
   \begin{minipage}{0.06\linewidth}
        \quad
     \end{minipage}%
   \begin{minipage}{0.42\linewidth}
     \centering
     \resizebox{1.0\linewidth}{!}{
     \ra{1.5}
        \begin{tabular}{rlccccc}
        \toprule
         &  & \multicolumn{2}{c}{R1} &  & \multicolumn{2}{c}{R5} \\
         &  & $|G_{norm}|$ & AUC &  & $|G_{norm}|$ & AUC \\
         \midrule
        surfer &  & 0.0812 & 0.6169 &  & 0.1441 & 0.6562 \\
        car24 &  & 0.5983 & 0.8299 &  & 0.6258 & 0.8312 \\
        blurCar &  & 0.2680 & 0.8125 &  & 0.4273 & 0.8141 \\
        box &  & 0.8377 & 0.5837 &  & 1.0286 & 0.7542 \\
        \bottomrule
        \end{tabular}
      }%
     \end{minipage}
    \hspace{0.05cm}
    \caption{\textbf{Proof of Concept.} Left: Convergence of the Gram determinant after repetitively re-running the tracker with THOR. Right: Gram determinant and area under curve evaluated (AUC) at the end of first and last runs (R$_1$ and R$_5$) of the experiment.}
    \label{fig:poc1}
    \vspace{-.3cm}
\end{figure}

To validate that the Gram determinant of the feature templates truly represents the diversity of information about the object, we build the following experiment. We run a tracker (SiamRPN) together with THOR on sequences from OTB2015 and observe the normalized Gram determinant $|G_{norm}|$ during the tracking. $G$ is normalized against $G_{11}$ to avoid numerical problems when calculating the determinant. At the end of a sequence, the final obtained templates are saved. We re-run the tracker while loading the previously saved templates and record the determinant again. We keep repeating this last step until the determinant converges. Surprisingly, the determinant does not stay constant after first reloading the templates. However, because the tracker yields different results with reloaded templates, the determinant keeps growing since it's exploring previously unseen candidate templates. Figure~\ref{fig:poc1} illustrates this behaviour. The convergence of the determinant represents the saturation of possible accumulated information from saved templates. Besides, we can observe that re-running the tracker with improved templates can also improve the AUC. The convergence of the determinant together with the improved AUC show that this measure truly enables the collection of good templates for tracking.

\subsection{State-of-the-Art Comparisons}\label{sec:sota_comparisons}
In this section, we determine the general performance regarding speed and the performance on the established visual tracking benchmarks VOT2018 and OTB2015.

\textbf{Performance on VOT2018.}
On VOT2018, performance is measured in terms of accuracy, robustness, and expected average overlap (EAO), where EAO is used to rank trackers. 
Table \ref{tab:benchmarks} shows that THOR is able to improve all state-of-the-art trackers in terms of EAO. THOR-SiamRPN even pushes the performance back to current SotA results of trackers with much more sophisticated network architectures such as SiamRPN++ \cite{li2019siamrpnpp}. SiamRPN++ achieves an EAO of 0.414 while running at 35 FPS (on a NVIDIA Titan X). THOR-SiamRPN (dynamic) achieves an EAO of 0.416 while running at 112 FPS.
The same strong improvements can be seen for robustness, which means that THOR mitigates tracking drift. 
Generally, the THOR enhanced trackers perform slighly worse on accuracy. 
In some sequences, the tracker puts up with a loss in accuracy in order to keep the object tracked, by predicting a larger bounding box. A second reason is THOR's disposition to track the entirety of an object (which it does by design), therefore in sequence with e.g. face-tracking only, THOR can start to track the entire head, not only the face. 

\textbf{Performance on OTB2015.}
On OTB2015, performance is measured with the area under curve (AUC) and the mean distance precision. As shown in Table \ref{tab:benchmarks} THOR improves all trackers on both metrics. Especially precision is improved by adding THOR to the trackers. Generally, both dynamic and ensemble lower bound yield similar results.

\textbf{Speed.} 
Table \ref{tab:benchmarks} shows that THOR slows the trackers down, which is to be expected of a multi-template tracker since there are necessarily additional computations. The general bigger decline for SiamFC can be explained with the expensive up-sizing operation in SiamFC, which is hard to parallelize in the given implementation. With a smaller and faster model, SiamRPN can reach speeds of $325$ FPS, whereas THOR-SiamRPN can run at a respectable speed of $244$ FPS. The experiments demonstrate that THOR still runs at a reasonable speed, especially when compared to other multi-template matching approaches \cite{lee2018memory,yang2018learning}. It also shows that if a tracker is using a faster model, THOR can also be run at higher speeds.

\begin{table}[htb!]
\centering

\ra{1.2}
\resizebox{1.0\linewidth}{!}{
\begin{tabular}{lcllccccllccc}
\toprule
 &  &  &  & \multicolumn{4}{c}{VOT2018} &  &  & \multicolumn{3}{c}{OTB2015} \\
Tracker & Lower Bound &  &  & Accuracy $\Uparrow$  & Robustness $\Downarrow$ & EAO $\Uparrow$ & Speed (FPS) $\Uparrow$ &  &  & AUC $\Uparrow$ & Precision $\Uparrow$ & Speed (FPS) $\Uparrow$ \\
\midrule
SiamFC & $-$ &  &  & $\textbf{0.5194}$ & $0.6696$ & $0.1955$ & $\textbf{219}$ &  &  & $0.5736$ & $0.6962$ & $\textbf{214}$ \\
THOR-SiamFC & dynamic &  &  & $0.4977$ & $0.4448$ & $0.2562$ & $99$ &  &  & $\textbf{0.5990}$ & $\textbf{0.7347}$ & $97$ \\
THOR-SiamFC & ensemble &  &  & $0.4846$ & $\textbf{0.3746}$ & $\textbf{0.2672}$ & $69$ &  &  & $0.5971$ & $0.7291$ & $80$ \\
\midrule
SiamRPN & $-$ &  &  & $\textbf{0.5858}$ & $0.3371$ & $0.3223$ & $\textbf{133}$ &  &  & $0.6335$ & $0.7674$ & $\textbf{137}$ \\
THOR-SiamRPN & dynamic &  &  & $0.5818$ & $0.2341$ & $\textbf{0.4160}$ & $112$ &  &  & $\textbf{0.6477}$ & $\textbf{0.7906}$ & $106$ \\
THOR-SiamRPN & ensemble &  &  & $0.5563$ & $\textbf{0.2248}$ & $0.3971$ & $105$ &  &  & $0.6407$ & $0.7867$ & $110$ \\
\midrule
SiamMask & $-$ &  &  & $\textbf{0.6096}$ & $0.2810$ & $0.3804$ & $\textbf{95}$ &  &  & $0.6204$ & $0.7683$ & $\textbf{97}$ \\
THOR-SiamMask & dynamic &  &  & $0.5891$ & $0.2388$ & $0.3846$ & $60$ &  &  & $\textbf{0.6397}$ & $0.7900$ & $78$ \\
THOR-SiamMask & ensemble &  &  & $0.5903$ & $\textbf{0.2013}$ & $\textbf{0.4104}$ & $70$ &  &  & $0.6319$ & $\textbf{0.7929}$ & $66$ \\
\bottomrule
\end{tabular}
}
\caption{\textbf{Tracking benchmarks.} The attained performances of the trackers on VOT2018 and OTB2015. The main metric for ranking trackers is EAO (expected average overlay) on VOT2018, and AUC (area under curve) on OTB2015.}
\label{tab:benchmarks}
\end{table}

\subsection{Ablation study}

\begin{figure}
\vspace{-0.5cm}
\centering
   \begin{minipage}{0.37\linewidth}
       \includegraphics[width=\textwidth]{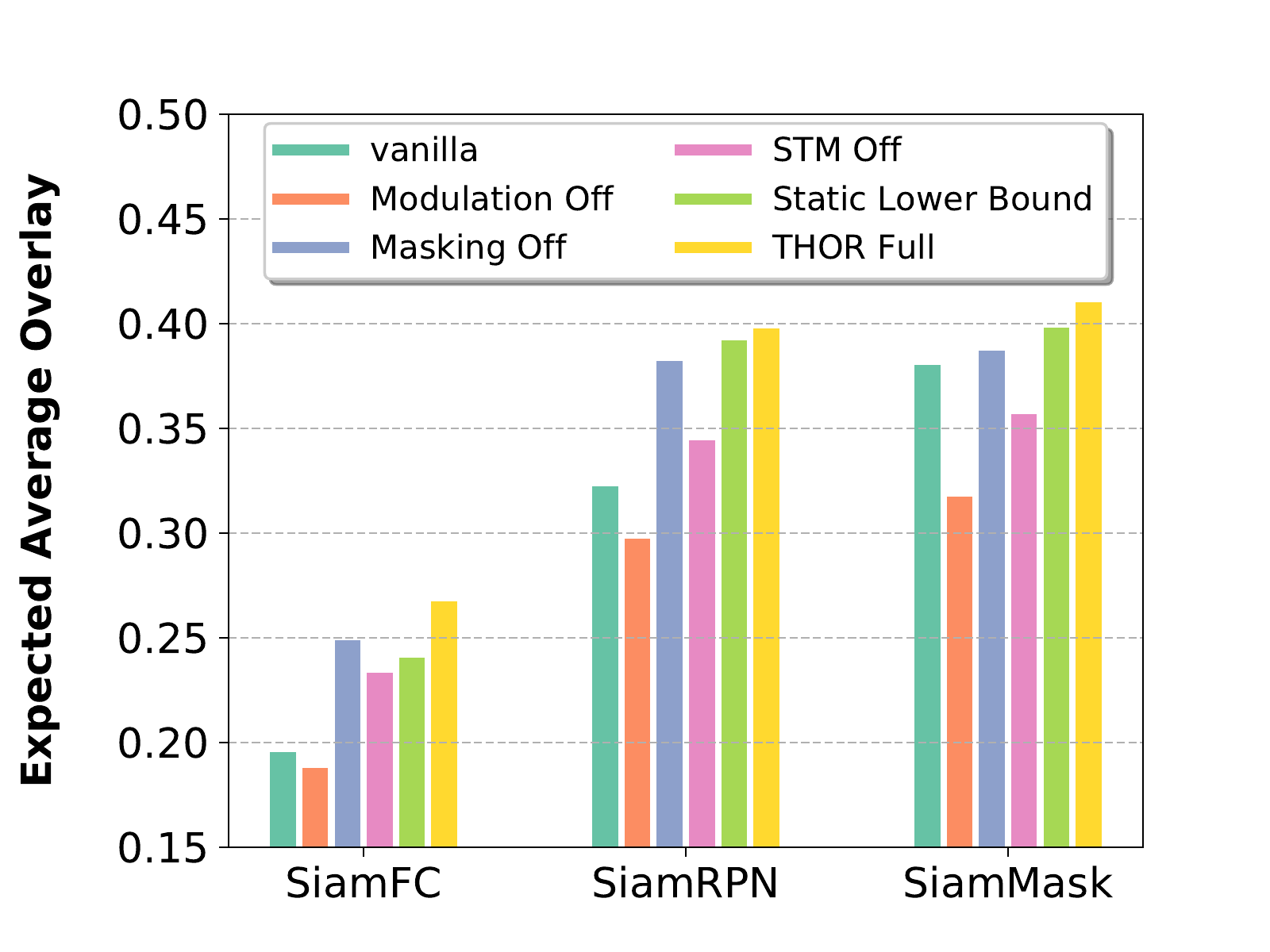}
     \end{minipage}%
    \begin{minipage}{0.06\linewidth}
        \quad
    \end{minipage}%
   \begin{minipage}{0.42\linewidth}
     \centering
     \resizebox{1.0\linewidth}{!}{
     \ra{1.3}
        \begin{tabular}{rcc}
        \toprule
         & \multicolumn{2}{c}{lower bound} \\
         & dynamic & ensemble \\
         \midrule
        Mean of $|G_{norm}|$ $\Uparrow$ & $0.0261$ & $\textbf{0.25164}$ \\
        $\#$ drifted templates $\Downarrow$& $\textbf{7}$ & $16$ \\
        $\#$ LT updates & $599$& $1797$ \\
        Relative drift $\Downarrow$& $1.17$ \% & $\textbf{0.89 \%}$ \\
        \bottomrule
        \end{tabular}
      }
     \end{minipage}
    \hspace{0.05cm}
    \caption{\textbf{Ablation Study.} Left: The effect on THOR's perfomance on VOT2018 when disabling modulation, masking, and the short-term module, or a static lower bound. Right:
    Comparison of the proposed strategies for the lower bound evaluated on OTB2015.
    }
    \label{fig:ablation}
\end{figure}

We introduced several concepts to enhance THOR's functionality. The LTM is improved by \textit{template masking} and a \textit{dynamic} or \textit{ensemble lower bound}. 
The \textit{STM} handles sequences with abrupt changes. 
We leverage the information of all templates through \textit{modulation}. In the following, we conduct experiments to determine the influence of these concepts.

\textbf{Tracking Performance.} To measure the impact of the concepts on the tracking performance, we conduct an ablation study on VOT2018. We compare the performance of all trackers with THOR and without ("vanilla"). Then we disable one of the concepts to determine their influence on the final performance.  Figure \ref{fig:ablation} (left) shows, that for all trackers the best performance (determined by EAO) can only be reached when all concepts are utilized. Turning off the modulation or the STM has the biggest negative impact on the final performance, which shows the importance of the respective concept. 

\textbf{Dynamic vs. Ensemble Lower Bound.}
To compare both proposed strategies for the lower bound, we record the normalized Gram determinant $|G_{norm}|$ at the end of every sequence in OTB2015. We then visually inspect the templates accumulated in the LTM and determine the number of drifted templates, i.e. when the tracked object is not in the center of the template. The relative drift is equal to the ratio of the number of drifted templates and the total number of updates. Figure \ref{fig:ablation} (right) shows that both strategies are effective in keeping the amount of drift low. However, the ensemble strategy manages to achieve a much higher mean of $|G_{norm}|$, indicating its ability to accumulate more diverse object representations (see also the qualitative comparison in the Appendix).

\section{Discussion and Future Work}
Although the presented framework demonstrates strong empirical performance, some aspects of it can still be improved. For instance, THOR's improvement over short sequences is not as great as for long ones.  This is expected since short sequences don't allow for accumulating enough templates. 
Furthermore, we only tested THOR on short-term tracking benchmarks. It would be interesting to observe how much improvement can be obtained for long-term tracking (for instance on OxUvA~\cite{valmadre2018long} or VOT-LT~\cite{VOT_TPAMI}). We speculate that THOR could have even more impact on long-term tracking. However, there is a higher risk of drift in long-term tracking, which needs to be addressed by a properly chosen lower bound. 
Moreover, siamese trackers are usually sensitive to the choice of hyperparameters which makes THOR similarly sensitive since it builds on top of them. This issue concerns the whole field of siamese trackers and should be addressed. 
Finally, the current version of THOR does not require any additional training besides the training of the tracker itself. Hence, a possible future direction would be to train the tracker to also optimize the THOR objective. We leave these improvements for future work.

\section{Conclusion}
We propose a framework for building holistic object representations for tracking. The presented approach, THOR, can be used together with any other template-matching tracker which uses inner products for similarity measuring. It accumulates short-term templates, representing sudden changes in objects appearances, and diverse long-term ones which allow representing objects in a more general way. 
Furthermore, we show that plugging our framework on top of state-of-the-art trackers improves their performance on tracking benchmarks without any need for further training of any parts of their networks. Moreover, the presented method barely reduces the speed of the original trackers. In future work, we plan on exploiting this holistic object representation and the presented diversity measure beyond visual tracking for active vision and robotic manipulation tasks.\\

\noindent
\textbf{Acknowledgements.} We gratefully acknowledge the general support of Microsoft Germany and the Alfried Krupp von Bohlen und Halbach Foundation.

\bibliography{thor}{}

\begin{thebibliography}{10}

\bibitem{bertinetto2016fully}
Luca Bertinetto, Jack Valmadre, Joao~F Henriques, Andrea Vedaldi, and Philip~HS
  Torr.
\newblock Fully-convolutional siamese networks for object tracking.
\newblock In {\em European conference on computer vision (ECCV)}, pages
  850--865. Springer, 2016.

\bibitem{black1998eigentracking}
Michael~J Black and Allan~D Jepson.
\newblock Eigentracking: Robust matching and tracking of articulated objects
  using a view-based representation.
\newblock {\em International Journal of Computer Vision (IJCV)}, 26(1):63--84,
  1998.

\bibitem{bolme2010visual}
David~S Bolme, J~Ross Beveridge, Bruce~A Draper, and Yui~Man Lui.
\newblock Visual object tracking using adaptive correlation filters.
\newblock In {\em Conference on Computer Vision and Pattern Recognition
  (CVPR)}, pages 2544--2550. IEEE, 2010.

\bibitem{held2016learning}
David Held, Sebastian Thrun, and Silvio Savarese.
\newblock Learning to track at 100 fps with deep regression networks.
\newblock In {\em European Conference on Computer Vision (ECCV)}, pages
  749--765. Springer, 2016.

\bibitem{hochreiter1997long}
Sepp Hochreiter and J{\"u}rgen Schmidhuber.
\newblock Long short-term memory.
\newblock {\em Neural computation}, 9(8):1735--1780, 1997.

\bibitem{hofhauser2008edge}
Andreas Hofhauser, Carsten Steger, and Nassir Navab.
\newblock Edge-based template matching and tracking for perspectively distorted
  planar objects.
\newblock In {\em International Symposium on Visual Computing}, pages 35--44.
  Springer, 2008.

\bibitem{hong2015multi}
Zhibin Hong, Zhe Chen, Chaohui Wang, Xue Mei, Danil Prokhorov, and Dacheng Tao.
\newblock Multi-store tracker (muster): A cognitive psychology inspired
  approach to object tracking.
\newblock In {\em Conference on Computer Vision and Pattern Recognition
  (CVPR)}, pages 749--758, 2015.

\bibitem{jurie2002real}
Fr{\'e}d{\'e}ric Jurie, Michel Dhome, et~al.
\newblock Real time robust template matching.
\newblock In {\em British Machine Vision Conference (BMVC)}, pages 1--10, 2002.

\bibitem{VOT_TPAMI}
Matej Kristan, Jiri Matas, Ale\v{s} Leonardis, Tomas Vojir, Roman Pflugfelder,
  Gustavo Fernandez, Georg Nebehay, Fatih Porikli, and Luka \v{C}ehovin.
\newblock A novel performance evaluation methodology for single-target
  trackers.
\newblock {\em IEEE Transactions on Pattern Analysis and Machine Intelligence
  (TPAMI)}, 38(11):2137--2155, Nov 2016.

\bibitem{lee2018memory}
Hankyeol Lee, Seokeon Choi, and Changick Kim.
\newblock A memory model based on the siamese network for long-term tracking.
\newblock In {\em European Conference on Computer Vision (ECCV)}, 2018.

\bibitem{li2019siamrpnpp}
Bo~Li, Wei Wu, Qiang Wang, Fangyi Zhang, Junliang Xing, and Junjie Yan.
\newblock Siamrpn++: Evolution of siamese visual tracking with very deep
  networks.
\newblock In {\em Conference on Computer Vision and Pattern Recognition
  (CVPR)}, pages 4282--4291, 2019.

\bibitem{li2018high}
Bo~Li, Junjie Yan, Wei Wu, Zheng Zhu, and Xiaolin Hu.
\newblock High performance visual tracking with siamese region proposal
  network.
\newblock In {\em Proceedings of the Conference on Computer Vision and Pattern
  Recognition (CVPR)}, pages 8971--8980. IEEE, 2018.

\bibitem{mohr2009continuous}
Daniel Mohr and Gabriel Zachmann.
\newblock Continuous edge gradient-based template matching for articulated
  objects.
\newblock In {\em Joint Conference on Computer Vision, Imaging and Computer
  Graphics Theory and Applications (VISIGRAPP)}, pages 519--524, 2009.

\bibitem{nguyen2001occlusion}
Hieu~Tat Nguyen, Marcel Worring, and Rein Van Den~Boomgaard.
\newblock Occlusion robust adaptive template tracking.
\newblock In {\em International Conference on Computer Vision (ICCV)},
  volume~1, pages 678--683. IEEE, 2001.

\bibitem{olson1997automatic}
Clark~F Olson and Daniel~P Huttenlocher.
\newblock Automatic target recognition by matching oriented edge pixels.
\newblock {\em IEEE Transactions on Image Processing}, 1997.

\bibitem{ren2015faster}
Shaoqing Ren, Kaiming He, Ross Girshick, and Jian Sun.
\newblock Faster r-cnn: Towards real-time object detection with region proposal
  networks.
\newblock In {\em Advances in Neural Information Processing Systems (NeurIPS)},
  pages 91--99, 2015.

\bibitem{steger2002occlusion}
Carsten Steger.
\newblock Occlusion, clutter, and illumination invariant object recognition.
\newblock {\em International Archives of Photogrammetry Remote Sensing and
  Spatial Information Sciences}, 34(3/A):345--350, 2002.

\bibitem{sukhbaatar2015end}
Sainbayar Sukhbaatar, Arthur Szlam, Jason Weston, and Rob Fergus.
\newblock End-to-end memory networks.
\newblock In {\em Advances in Neural Information Processing Systems (NeurIPS)},
  pages 2440--2448, 2015.

\bibitem{tao2016siamese}
Ran Tao, Efstratios Gavves, and Arnold~WM Smeulders.
\newblock Siamese instance search for tracking.
\newblock In {\em Conference on Computer Vision and Pattern Recognition
  (CVPR)}, pages 1420--1429, 2016.

\bibitem{valmadre2017end}
Jack Valmadre, Luca Bertinetto, Jo{\~a}o Henriques, Andrea Vedaldi, and
  Philip~HS Torr.
\newblock End-to-end representation learning for correlation filter based
  tracking.
\newblock In {\em Proceedings of the IEEE Conference on Computer Vision and
  Pattern Recognition (CVPR)}, pages 2805--2813, 2017.

\bibitem{valmadre2018long}
Jack Valmadre, Luca Bertinetto, Joao~F Henriques, Ran Tao, Andrea Vedaldi,
  Arnold~WM Smeulders, Philip~HS Torr, and Efstratios Gavves.
\newblock Long-term tracking in the wild: A benchmark.
\newblock In {\em European Conference on Computer Vision (ECCV)}, pages
  670--685, 2018.

\bibitem{wang2018fast}
Qiang Wang, Li~Zhang, Luca Bertinetto, Weiming Hu, and Philip~HS Torr.
\newblock Fast online object tracking and segmentation: A unifying approach.
\newblock {\em arXiv preprint arXiv:1812.05050}, 2018.

\bibitem{weston2014memory}
Jason Weston, Sumit Chopra, and Antoine Bordes.
\newblock Memory networks.
\newblock {\em arXiv preprint arXiv:1410.3916}, 2014.

\bibitem{OTB2015}
Y.~{Wu}, J.~{Lim}, and M.~{Yang}.
\newblock Object tracking benchmark.
\newblock {\em IEEE Transactions on Pattern Analysis and Machine Intelligence
  (TPAMI)}, 37(9):1834--1848, 2015.

\bibitem{yang2018learning}
Tianyu Yang and Antoni~B Chan.
\newblock Learning dynamic memory networks for object tracking.
\newblock In {\em European Conference on Computer Vision (ECCV)}, pages
  152--167, 2018.

\bibitem{zhu2018distractor}
Zheng Zhu, Qiang Wang, Bo~Li, Wei Wu, Junjie Yan, and Weiming Hu.
\newblock Distractor-aware siamese networks for visual object tracking.
\newblock In {\em European Conference on Computer Vision (ECCV)}, pages
  101--117, 2018.

\end{thebibliography}
\bibliographystyle{plain}
\pagebreak
\appendix
\section{Evolution of LTM slots over time}

From left to right: template at the beginning, in the middle and at the end of the sequence. The sequences are from OTB2015 (Girl, Singer1, Toy, Vase) and we use SiamRPN for tracking. \ref{subsection:dyn} shows the templates using the dynamic lower bound, \ref{subsection:en} the ensemble lower bound.

\subsection{Dynamic Lower Bound}\label{subsection:dyn}
\begin{figure}[!htb]
    \centering
    \begin{subfigure}{0.3\textwidth}
        \centering
        \includegraphics[width=\linewidth]{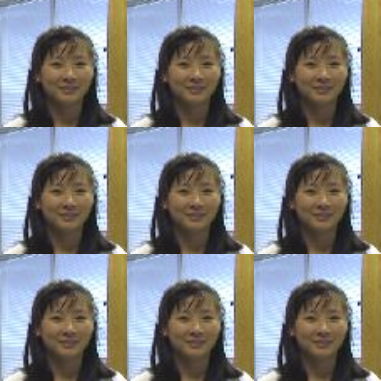}
    \end{subfigure}
    \begin{subfigure}{0.3\textwidth}
        \centering
        \includegraphics[width=\linewidth]{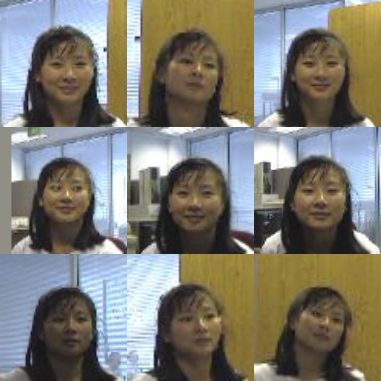}
    \end{subfigure}
    \begin{subfigure}{0.3\textwidth}
        \centering
        \includegraphics[width=\linewidth]{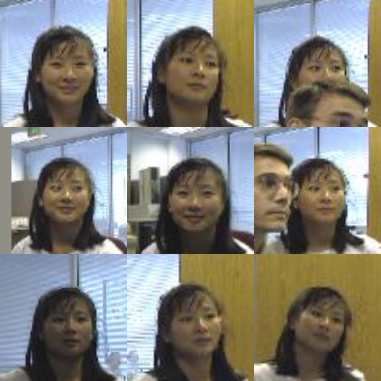}
    \end{subfigure}
\end{figure}

\vspace{-0.7cm}

\begin{figure}[!htb]
    \centering
    \begin{subfigure}{0.3\textwidth}
        \centering
        \includegraphics[width=\linewidth]{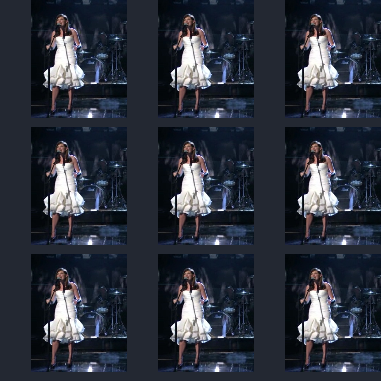}
    \end{subfigure}
    \begin{subfigure}{0.3\textwidth}
        \centering
        \includegraphics[width=\linewidth]{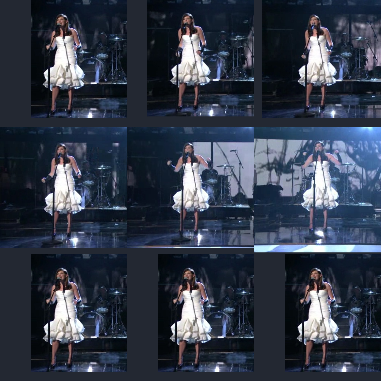}
    \end{subfigure}
    \begin{subfigure}{0.3\textwidth}
        \centering
        \includegraphics[width=\linewidth]{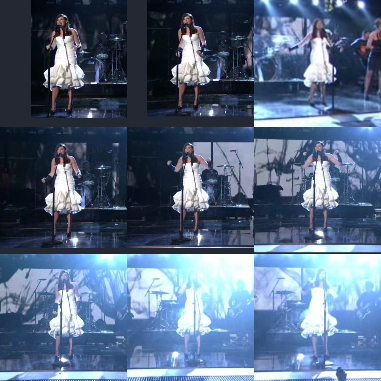}
    \end{subfigure}
\end{figure}

\vspace{-0.7cm}

\begin{figure}[!htb]
    \centering
    \begin{subfigure}{0.3\textwidth}
        \centering
        \includegraphics[width=\linewidth]{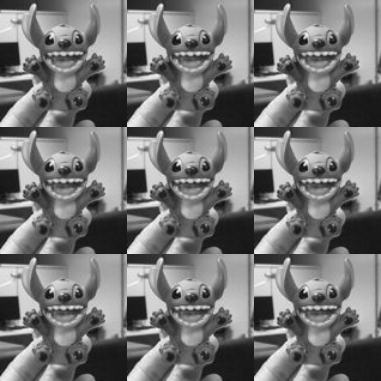}
    \end{subfigure}
    \begin{subfigure}{0.3\textwidth}
        \centering
        \includegraphics[width=\linewidth]{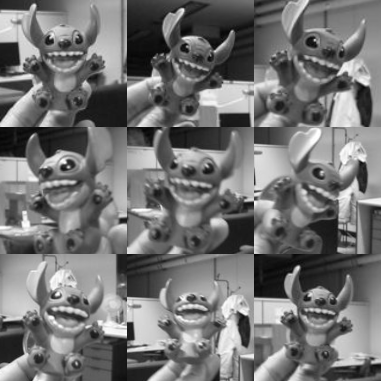}
    \end{subfigure}
    \begin{subfigure}{0.3\textwidth}
        \centering
        \includegraphics[width=\linewidth]{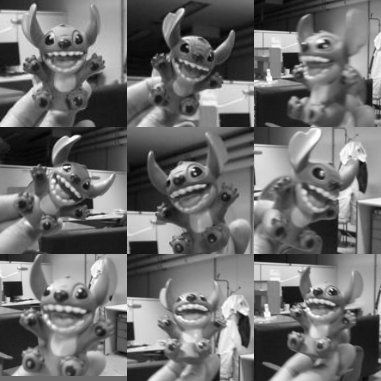}
    \end{subfigure}
\end{figure}

\vspace{-0.7cm}

\begin{figure}[!htb]
    \centering
    \begin{subfigure}{0.3\textwidth}
        \centering
        \includegraphics[width=\linewidth]{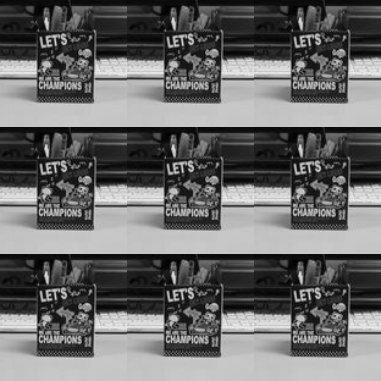}
    \end{subfigure}
    \begin{subfigure}{0.3\textwidth}
        \centering
        \includegraphics[width=\linewidth]{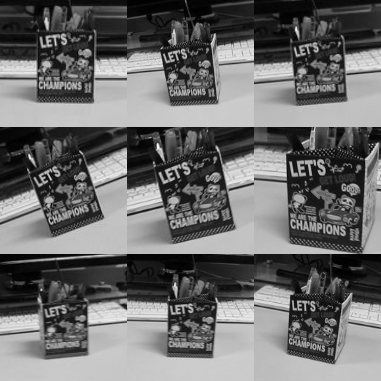}
    \end{subfigure}
    \begin{subfigure}{0.3\textwidth}
        \centering
        \includegraphics[width=\linewidth]{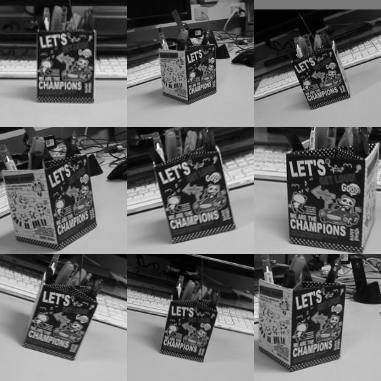}
    \end{subfigure}
\end{figure}

\subsection{Ensemble-based Lower Bound}\label{subsection:en}
\begin{figure}[!htb]
    \centering
    \begin{subfigure}{0.3\textwidth}
        \centering
        \includegraphics[width=\linewidth]{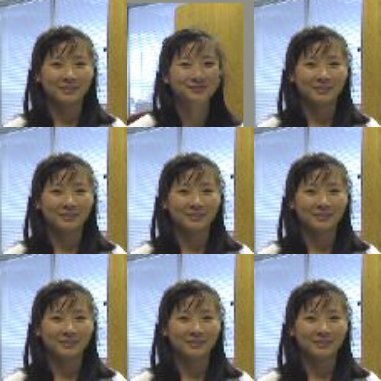}
    \end{subfigure}
    \begin{subfigure}{0.3\textwidth}
        \centering
        \includegraphics[width=\linewidth]{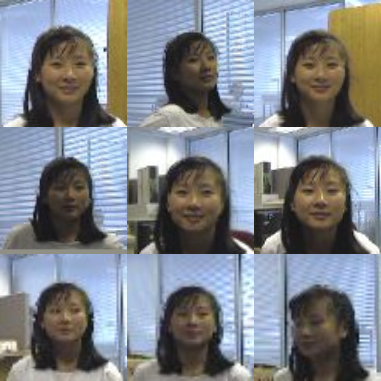}
    \end{subfigure}
    \begin{subfigure}{0.3\textwidth}
        \centering
        \includegraphics[width=\linewidth]{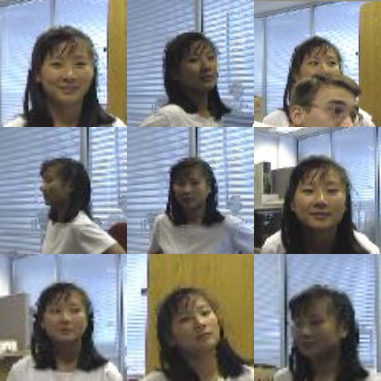}
    \end{subfigure}
\end{figure}

\vspace{-0.7cm}

\begin{figure}[!htb]
    \centering
    \begin{subfigure}{0.3\textwidth}
        \centering
        \includegraphics[width=\linewidth]{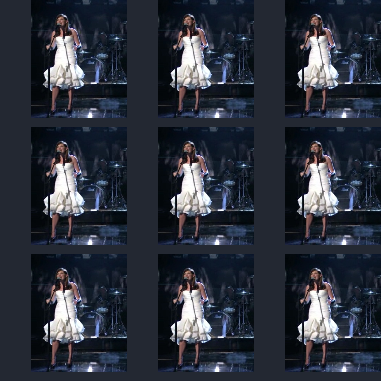}
    \end{subfigure}
    \begin{subfigure}{0.3\textwidth}
        \centering
        \includegraphics[width=\linewidth]{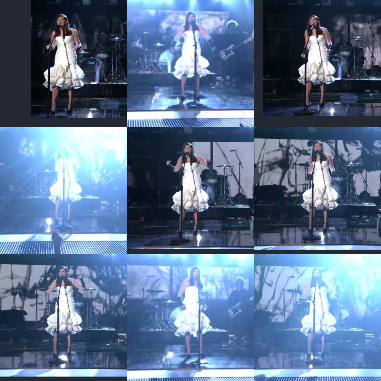}
    \end{subfigure}
    \begin{subfigure}{0.3\textwidth}
        \centering
        \includegraphics[width=\linewidth]{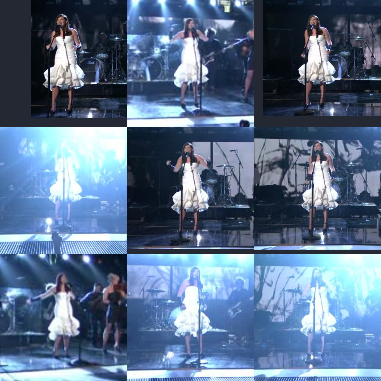}
    \end{subfigure}
\end{figure}

\vspace{-0.7cm}

\begin{figure}[!htb]
    \centering
    \begin{subfigure}{0.3\textwidth}
        \centering
        \includegraphics[width=\linewidth]{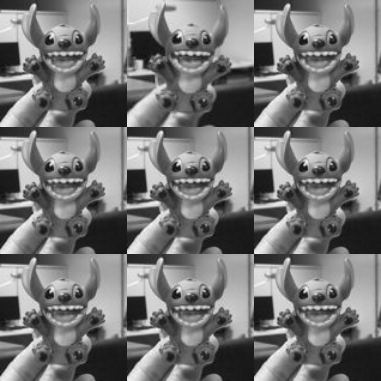}
    \end{subfigure}
    \begin{subfigure}{0.3\textwidth}
        \centering
        \includegraphics[width=\linewidth]{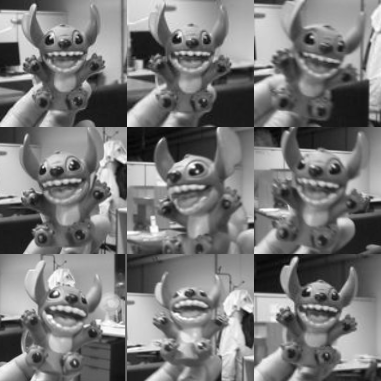}
    \end{subfigure}
    \begin{subfigure}{0.3\textwidth}
        \centering
        \includegraphics[width=\linewidth]{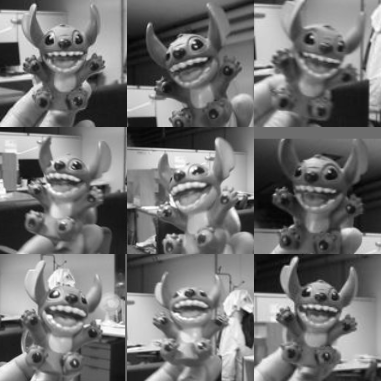}
    \end{subfigure}
\end{figure}

\vspace{-0.7cm}

\begin{figure}[!htb]
    \centering
    \begin{subfigure}{0.3\textwidth}
        \centering
        \includegraphics[width=\linewidth]{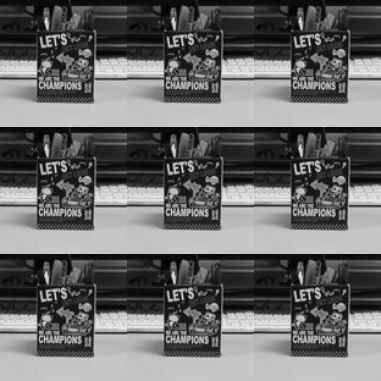}
    \end{subfigure}
    \begin{subfigure}{0.3\textwidth}
        \centering
        \includegraphics[width=\linewidth]{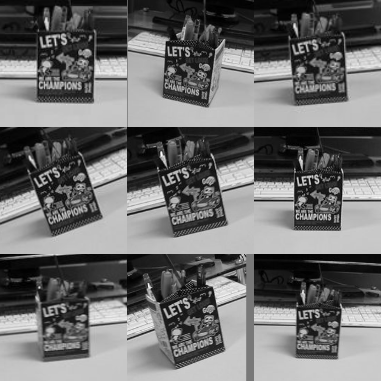}
    \end{subfigure}
    \begin{subfigure}{0.3\textwidth}
        \centering
        \includegraphics[width=\linewidth]{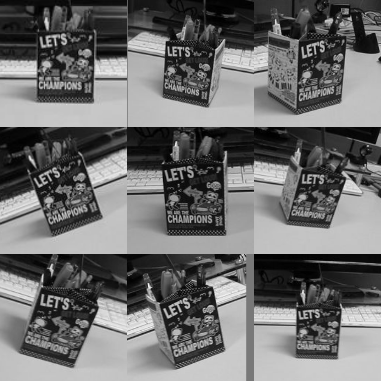}
    \end{subfigure}
\end{figure}
\end{document}